\PassOptionsToPackage{dvipsnames}{xcolor}
\documentclass{optica-article}

\journal{opticajournal} 

\articletype{Research Article}

\usepackage{lineno}

\definecolor{blue}{rgb}{0.21,0.49,0.74}
\definecolor{red}{rgb}{0.89,0.10,0.11}
\usepackage{graphicx}
\usepackage{amsmath}
\usepackage{booktabs}
\usepackage{float}
\usepackage{stfloats}
\usepackage{tikz}
\usetikzlibrary{spy,backgrounds}
\usepackage{diagbox}
\usepackage{multirow}
\usepackage{wrapfig}
\usepackage{lipsum}
\usepackage{color, colortbl}
\usepackage{caption}
\usepackage{pifont}
\usepackage{enumitem}
\usepackage{adjustbox}

\newcommand{\redcross}{\textcolor{red}{\ding{55}}}

\usepackage{xspace}

\usepackage{makecell}   

\begin{document}

\title{Task-Driven Lens Design}

\author{Xinge Yang,\authormark{1} Qiang Fu,\authormark{1} Yunfeng Nie,\authormark{2} and Wolfgang Heidrich\authormark{1,*}}

\address{\authormark{1}King Abdullah University of Science and Technology, Thuwal, 23955-6900, Saudi Arabia\\
\authormark{2}Vrije Universiteit Brussel (VUB), Brussels Photonics Team, Department of Applied Physics and Photonics, Pleinlaan 2, B-1050 Brussels, Belgium
}

\email{\authormark{*}wolfgang.heidrich@kaust.edu.sa} 


\begin{abstract}
Classical lens design minimizes optical aberrations to produce sharp images, but is typically decoupled from downstream computer vision tasks. Existing end-to-end optical design learns optical encoding through joint optimization, but often suffers from an unstable training process. We propose task-driven lens design, a new optimization philosophy for joint optics-network systems. We freeze the pretrained vision model and optimize only the lens so that the image formation better fits the model's feature preferences. This network-frozen setting yields a low-dimensional and stable optimization process, enabling lens design from scratch without human intervention, thereby exploring a broader design space. Multiple computer vision experiments show that TaskLenses outperform classical ImagingLenses with the same or even fewer elements. Our analysis reveals that the learned optics exhibit long-tailed point spread functions, better preserving preferred structural cues when aberrations cannot be fully corrected. These results highlight task-driven design as a practical route for optical lenses that are compatible with modern vision models, and also inspire new optical design objectives beyond traditional aberration minimization.
\end{abstract}
\section{Introduction}

Classical lens design is typically decoupled from any downstream image analysis task and focuses on maximizing image quality by minimizing optical aberrations. To achieve optimal performance for modern computer vision networks, such as those used for image classification~\cite{he2016deep, liu2022swin}, object detection~\cite{redmon2016you, zhang2022dino}, semantic segmentation~\cite{liu2021swin, zhai2022scaling}, and vision-language models (VLMs)~\cite{radford2021learning, li2022blip, zhang2024vision}, optical engineers design lenses that minimize aberrations to capture sharp, high-quality images. However, these high-end lenses often come at a prohibitively high cost and possess a bulky form factor with complex structures. For instance, modern smartphone cameras may include more than five highly aspheric elements~\cite{shabtay2020folded, zhou2023camera}, while professional camera lenses can incorporate six or more precision-manufactured elements~\cite{abe2018zoom, sugita2015zoom} to achieve satisfactory optical performance. For deployment into mobile and robotic platforms, it is desirable to reduce this optical complexity without compromising the performance of any downstream task. However, when residual aberrations persist in captured images, downstream computer vision performance can suffer dramatically.

Recent studies in end-to-end lens design~\cite{sitzmann2018end, sun2021end, tseng2021differentiable, wang2022differentiable} have demonstrated the potential for achieving optimal system-level performance by jointly optimizing camera optics and computer vision algorithms. This approach enables the learning of optical encoding power during the joint optimization of lenses and neural networks, yielding promising results across a variety of tasks, including compact imaging systems~\cite{jiang2024minimalist, na2024end, tseng2021neural}, extended depth-of-field imaging~\cite{sun2021end, wang2022differentiable, li2021end, yang2023curriculum, liu2021end, yang2024end}, hyperspectral imaging~\cite{jeon2019compact, dun2020learned, baek2021single, li2022quantization}, high-dynamic-range imaging~\cite{metzler2020deep, sun2020learning}, object detection~\cite{tseng2021differentiable, cote2022differentiable}, and depth estimation~\cite{chang2019deep, ikoma2021depth}. These works suggest that producing conventionally sharp images is often suboptimal for downstream tasks, especially when using simple optics where aberrations cannot be fully corrected.

However, learning optical encoding power through joint optimization of optics and neural networks presents several significant challenges. Optical systems typically involve optimizing only tens of physical parameters, whereas modern neural networks can contain millions to billions of parameters; optimizing them together often leads to unstable training dynamics and optimization oscillations. To achieve promising convergence, existing studies typically rely on pre-optimized lenses as starting points~\cite{sun2021end, tseng2021differentiable, wang2022differentiable, cote2022differentiable, cote2021deep}, which may trap the design process in local minima and limit exploration of the optical design space. Moreover, modern vision foundation models have evolved to encapsulate rich prior knowledge of the physical world; retraining or fine-tuning these large-scale networks is not only prohibitively expensive but also risks disrupting their robust, pre-learned representations. Prior works attempt to explore the possibility of simplifying optical systems according to computer vision requirements~\cite{cote2022differentiable, teh2025automated}; however, a comprehensive evaluation across diverse tasks and network architectures, as well as an explanation of the underlying optical mechanisms behind performance gains, remains missing.

In this paper, we introduce a new optimization philosophy for optics-network joint systems: freeze the network and optimize only the lens (i.e., \textbf{task-driven lens design}). While we use a differentiable image formation model (as in prior end-to-end works) to propagate gradients to the optical parameters, our network-frozen setting turns lens design into a low-dimensional and stable optimization problem. During the optimization, the lens learns to encode the image features preferred by the downstream computer vision models, which not only reduces optimization difficulty, also represents an explainable optical design objective.

We first demonstrate our task-driven lens design using image classification. We designed three distinct lenses (``TaskLenses'') fully automatically from scratch, using a frozen image classification network as the optimization objective. This automated task-driven lens design not only demonstrates the efficacy of stable network-provided gradients, but also enables a broader search space for lens design. For comparison, we designed three corresponding classical lenses (``ImagingLenses'') for each ``TaskLens'' by minimizing optical aberrations. We evaluated the image classification accuracies on the ImageNet benchmark~\cite{deng2009imagenet} and found that our ``TaskLenses'' consistently outperform the ``ImagingLenses'' in classification accuracy while utilizing the same number of lens elements. Moreover, we discovered that ``TaskLenses'' can achieve comparable, or even superior, image classification accuracy with fewer elements.

We further investigate the generalizability of the task-driven design across different computer vision tasks, including object detection, semantic segmentation, and vision-language models. By evaluating the task-driven designed lenses across various applications, we found that different computer vision tasks can share similar preferable image features, making it feasible to generalize from simpler to more complex tasks. We also observed the potential for generalization from simpler network architectures to more complex ones. Analysis of the optical characteristics of the designed lenses reveals that the TaskLenses tend to converge on long-tailed point spread functions (PSFs) with a sharp central peak, which is a significant deviation from the compact PSFs with a relatively broad central peak, obtained with conventional lens design. While such PSF profiles may not appear sharp or meet traditional optical design standards, it is more effective at preserving crucial high-frequency features against optical aberrations, thereby benefiting various computer vision tasks. The contributions of this paper are summarized as follows:
\begin{itemize}
    \item We introduce a new optimization philosophy for joint optics-network systems by freezing a pretrained vision model and optimizing only the lens. This task-driven design represents an explainable objective that is aligned with modern computer vision feature extraction.
    \item Task-driven lens design reduces optimization difficulty and enables broader exploration of the design space by starting from scratch, showing the capability of designing simplified structures with comparable computer vision performance using fewer lens elements.
    \item We provide a comprehensive evaluation of the task-driven designed lenses across various computer vision tasks and network architectures, demonstrating their generalizability and the emergence of a novel long-tailed PSF as a key characteristic of computer vision-preferred optical properties.
\end{itemize}
\begin{figure*}[t]
    \centering
    \includegraphics[width=\textwidth]{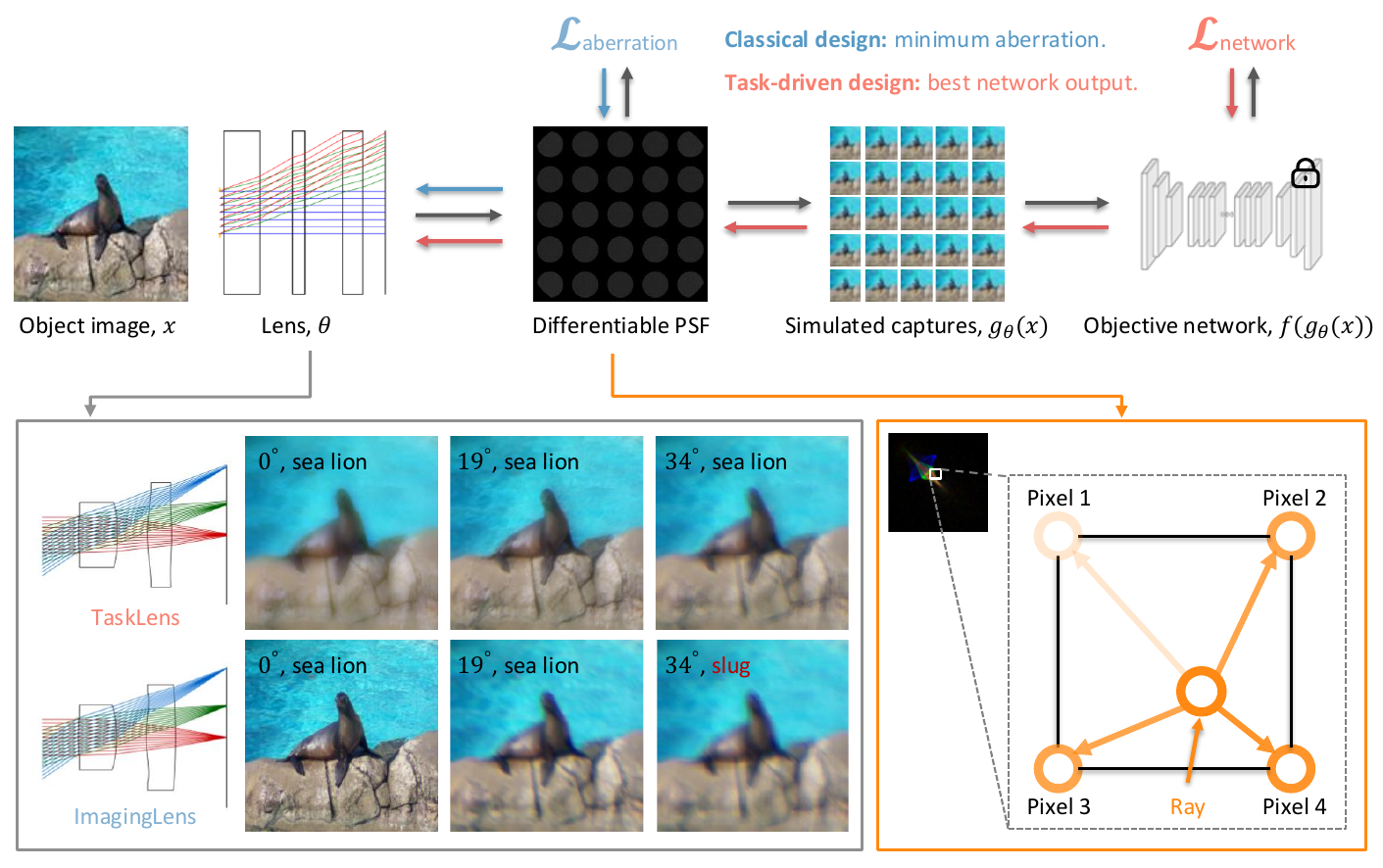}
    \caption{The task-driven lens design pipeline begins by calculating the PSF via differentiable ray tracing. This PSF is then convolved with input images to simulate camera captures. A well-trained computer vision network (e.g., for image classification) evaluates the final output error and backpropagates the loss function to optimize lens parameters. Unlike classical lens design, which minimizes optical aberrations, task-driven lens design prioritizes capturing image features that are most effective for downstream computer vision networks. After optimization, the designed TaskLens better preserves image structure details even in the presence of optical aberrations, leading to improved performance on the target computer vision task. In contrast, an ImagingLens, designed to produce clear captures, is more susceptible to performance degradation from optical aberrations, potentially resulting in misclassification.}
    \label{fig:pipeline}
\end{figure*}

\section{Methods}

\subsection{Task-driven lens design}
Lens design for a computer vision task can be formulated as:
\begin{equation}
    \theta^* = \underset{\theta}{\operatorname{argmin}}
    \left\|f_\phi(g_{\theta}(x)) - y\right\|,
    \label{eq:lens_design}
\end{equation}
In Equation~\eqref{eq:lens_design}, $f_\phi$ is a pre-trained neural network for the vision task:
\begin{equation}
    f_\phi = \underset{\phi}{\operatorname{argmin}} \left\|f(x) - y\right\|.
\end{equation}
Here, $x$ represents the object image, $\theta$ and $\phi$ respectively denote the parameters of the lens design and the neural network, $g$ represents the imaging process, $y$ is the ground truth label for the computer vision task $f_\phi$, and $f_\phi$ is a pre-trained computer vision model trained on a large dataset.

Classical lens design typically focuses on minimizing optical aberrations, such as root mean square (RMS) spot size or wavefront error. Under this design philosophy, optical engineers assume that minimizing optical aberrations to achieve optimal image quality will yield the best possible performance for downstream computer vision tasks. This classical lens design approach is illustrated by the blue arrow in Figure~\ref{fig:pipeline} and can be expressed as:
\begin{equation}
    \theta^* = \underset{\theta}{\operatorname{argmin}}\ h_{\theta}, 
    \label{eq:classical}
\end{equation}
where $h_{\theta}$ represents the optical aberration function, which depends on the lens parameters $\theta$. Ideally, when $h_{\theta} = 0$, the camera-captured image $g_{\theta}(x)$ is identical to the object image $x$ (i.e., $g_{\theta}(x) = x$). However, the classical lens design objective is decoupled from computer vision, and Equation~\eqref{eq:classical} represents an {\em intermediate} optimization objective for the target computer vision task. The mismatch between minimizing optical aberrations and the actual requirements of diverse computer vision applications leads to a performance gap. Consequently, when optical aberrations cannot be fully corrected due to constraints on the optical form factor or manufacturing limitations, especially in edge devices such as robotics, the corresponding computer vision performance can be significantly affected.

A more straightforward and promising lens design objective is to directly optimize the objective function in Equation~\eqref{eq:lens_design}, which can be expressed as:
\begin{equation}
    \theta^* = \underset{\theta}{\operatorname{argmin}} \left\|f(g_{\theta}(x)) - y\right\|.
    \label{eq:task_driven}
\end{equation}
This task-driven lens design objective directly optimizes the lens design for optimal computer vision performance on the target task, as illustrated by the red arrow in Figure~\ref{fig:pipeline}. During optimization, this objective evaluates how well the captured image is suited for the downstream computer vision network. However, realizing this task-driven lens design objective has been challenging in classical lens design due to the lack of a differentiable lens simulator $g_{\theta}(x)$, which is necessary to enable direct gradient backpropagation from the network $f$ outputs to the lens parameters $\theta$. Recent advances in differentiable optics have provided a new approach for task-driven lens design, as demonstrated in~\cite{wang2022differentiable, yang2023curriculum} and detailed in Section~\ref{sec:diff_psf}.

Modern vision foundation models trained on massive image datasets learn the rich information of image data distributions and can effectively extract useful latent features from image inputs. This inspires lens design to focus on capturing the most important image features from the objects, especially when the optical structure is limited by form factor and manufacturing constraints. Decomposing the original object image $x$ as $x = x_{f} \oplus x_{bg}$, where $x_{f}$ denotes the latent image features preferred by vision tasks and $x_{bg}$ denotes background information, combined via the operator $\oplus$, the task-driven lens design objective can further transform the imaging problem into a feature encoding problem, with the preferable image features $x_{f}$ defined by the computer vision models. Importantly, modern computer vision models are usually trained with massive image datasets that include various image degradations as data augmentation, thereby preventing the lens design from converging to that of a perfect imaging lens.


\subsection{Differentiable point spread function}
\label{sec:diff_psf}

The point spread function (PSF) characterizes the blurring effect of an optical system on a point light source. In image simulation, convolution of the PSF with the object image allows for simulating the image formation process of an optical lens. To compute the PSF, we trace optical rays emanating from point sources in the object space to the sensor plane. As illustrated in Figure~\ref{fig:pipeline}, the PSF is then computed by integrating the energy deposited by the ray intersections over the sensor pixels.  This process can be mathematically expressed as:
\begin{equation}
  \mathrm{PSF}(\mathbf{o}_p) = \sum_{i=1}^{N} u_i \, 
  \sigma\!\left(\frac{|(\mathbf{o}_p - \mathbf{o}_i) \cdot \hat{\mathbf{e}}_x|}{L}\right) 
  \sigma\!\left(\frac{|(\mathbf{o}_p - \mathbf{o}_i) \cdot \hat{\mathbf{e}}_y|}{L}\right),
  \label{eq:diff_psf}
\end{equation}
where $\mathbf{o}_p$ represents the pixel coordinate, $\mathbf{o}_i$ is the intersection point of the $i$-th ray with the sensor plane, and $N$ denotes the number of rays traced from the point source. The scalar $u_i$ signifies the radiant energy carried by the $i$-th ray, assigned a value of 1 in the experiments conducted. The vectors $\hat{\mathbf{e}}_x$ and $\hat{\mathbf{e}}_y$ are unit vectors situated within the sensor plane, and $L$ denotes the physical width of a sensor pixel. The weighting function $\sigma(\cdot)$ is defined as:
\begin{equation}
  \sigma(x) =
  \begin{cases}
    1 - x, & 0 \leq x \leq 1\\
    0, & \text{otherwise}
  \end{cases},
\end{equation}
which quantifies the contribution of a ray to its neighboring pixels. As depicted in Figure~\ref{fig:pipeline}, the energy of each ray is distributed among its four nearest neighboring pixels. Thus, Equation~\eqref{eq:diff_psf} can be interpreted as an instance of inverse bilinear interpolation. By leveraging this sub-pixel information, we can approximate the continuous light distribution using a finite number of rays. Since Equation~\eqref{eq:diff_psf} is differentiable, we can compute the gradient of the PSF concerning the ray positions. During the backpropagation phase, these gradients direct the rays toward the target pixels, facilitating updates through the optical rays to the lens surfaces, which in turn adjusts the lens parameters.

\section{Results}

In this section, we first assess the effectiveness of the task-driven lens design objective for optimizing image classification lenses from scratch (Sections~\ref{sec:details} and~\ref{sec:img_classi}). We demonstrate that lenses designed with specific computer vision objectives can achieve competitive, or even superior, image classification accuracy with fewer optical elements compared to classical lens designs (Section~\ref{sec:img_classi}). Subsequently, we extend this task-driven design approach to a variety of downstream computer vision tasks, including object detection, semantic segmentation, and vision-language models. We examine the underlying similarities across different computer vision tasks (Section~\ref{sec:downstream_tasks}) and their preferences regarding optical feature encoding, demonstrating the potential to design task-specific lenses for simpler tasks that generalize to more complex computer vision tasks.

\subsection{Implementation details}
\label{sec:details}

We conduct differentiable ray tracing for optical simulation using the open-source simulator \textsc{DeepLens}~\cite{wang2022differentiable,yang2023curriculum}. Each lens surface is modeled as an asphere, with curvature, axial position, and fourth- to tenth-order polynomial coefficients ($\alpha_{4}$ to $\alpha_{10}$) treated as optimizable parameters. For lens parameter optimization, we employ the Adam optimizer~\cite{loshchilov2017decoupled}, setting the learning rate to $1 \times 10^{-4}$ for curvature, position, and $\alpha_{4}$ coefficients. An exponential decay factor of 0.02 is applied to the learning rates of the higher-order coefficients. To prevent self-intersections, we impose penalties on the distances between lens surfaces. The lens materials are selected from a library of common smartphone-grade polymers. The image sensor has a resolution of $1080 \times 1920$ pixels and a pixel pitch of $1.8 \, \mu\text{m}$. For the calculation of the PSF, we trace rays at three wavelengths: 656.3 nm, 589.3 nm, and 486.1 nm. More implementation settings are detailed in the supplementary material. Instead of simulating full sensor resolution images, we consider low-resolution image patches for different image regions covering the full field of view (FoV). This choice stems from the fact that modern computer vision models are typically trained on low-resolution images (e.g., $224 \times 224$), even though camera sensors often possess megapixel resolutions (e.g., 12 MP). This discrepancy in resolution necessitates simulating low-resolution images for task-driven lens design. During training, we randomly sample FoVs and, for each FoV, we trace 4096 rays and compute a $101 \times 101$ PSF kernel. Starting from a random initialization, the lens is trained for approximately 3,000 iterations, with the downstream network held constant. For evaluation, we select nine FoVs to represent the lens's performance.


\subsection{Task-driven lens design for image classification}
\label{sec:img_classi}

\begin{figure*}[!htp]
    \centering
    \includegraphics[width=\textwidth]{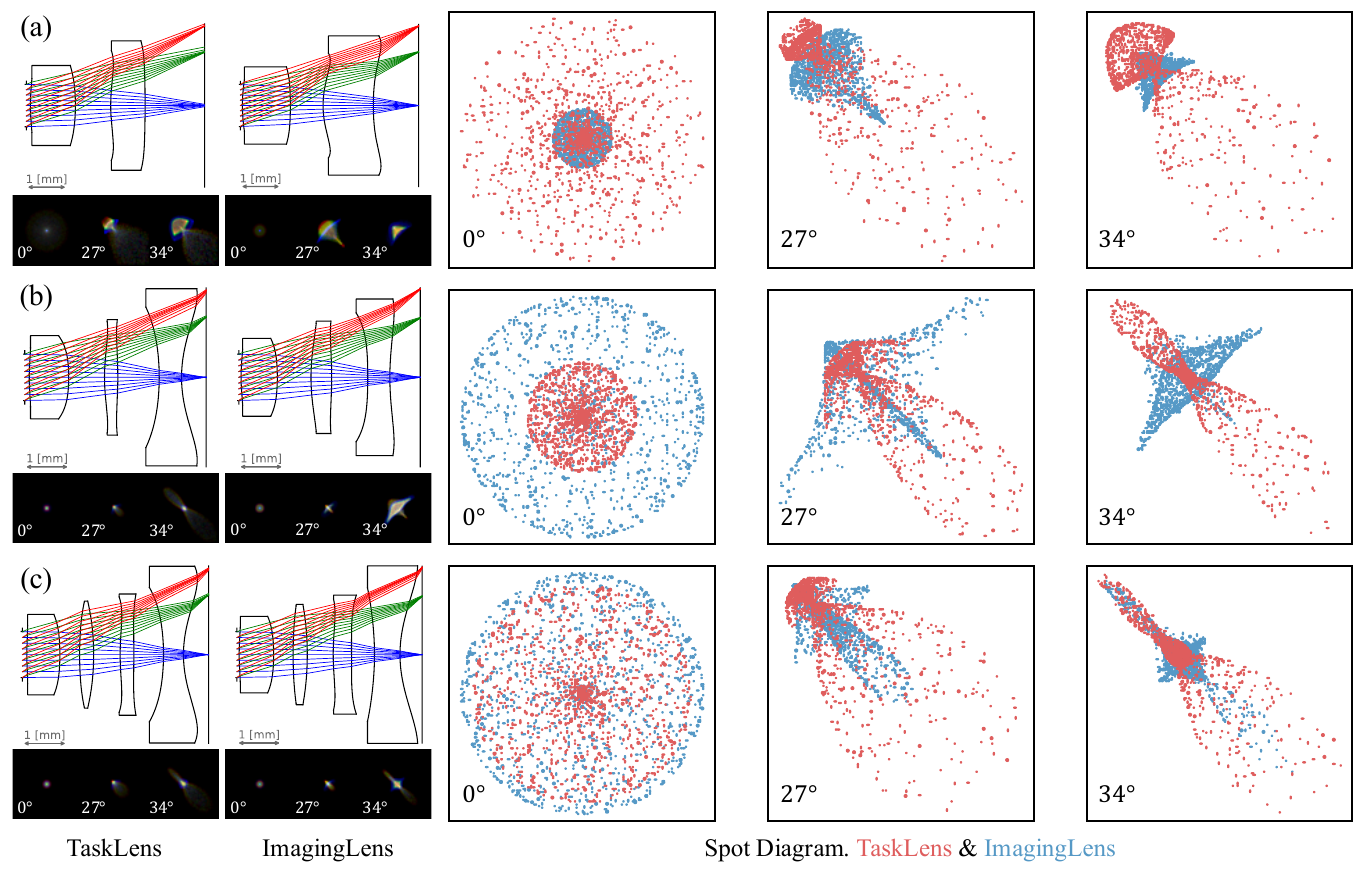}
    \caption{The lens layout, PSFs and corresponding spot diagrams at different fields are shown for 2P (a), 3P (b), and 4P (c) lenses. Although RMS spot size of the TaskLens at some fields is larger than that of the ImagingLens, the majority of optical rays always converge within a small region. This spot diagram distribution results in a long-tailed PSF, which is characterized by a small, concentrated center and sparsely populated outer regions. In scenarios where the optical system cannot fully correct all aberrations, this long-tailed PSF with a sharp central peak is effective in preserving essential features of the object images, which benefits the performance of computer vision models.}
    \label{fig:detailed_comparison}
\end{figure*}

We first demonstrate the effectiveness of task-driven lens design by designing lenses for image classification, a fundamental task in computer vision that is essential for image understanding and feature encoding. Utilizing a well-trained ResNet-50 network~\cite{he2016deep}, we automatically design three distinct lenses comprising two, three, and four elements, entirely without human intervention (Figure~\ref{fig:detailed_comparison}). All three lens designs achieve a FoV of $68.8^{\circ}$ and an F-number of 2.8. This success validates the effectiveness of employing gradients derived from network outputs and the robustness of the proposed task-driven lens design methodology. By initiating design from scratch and employing automated optimization, we circumvent the potential local minima associated with human expertise, thereby exploring a broader design space that extends beyond classical lens design paradigms.

We compare each of our task-driven lenses, referred to as ``TaskLens'', against three lenses designed with classical objectives focused on minimizing aberrations, labeled as ``ImagingLens''. The first ImagingLens (\#1) is optimized using the open-source \textsc{DeepLens} code~\cite{yang2023curriculum} from scratch by minimizing the root-mean-square (RMS) spot size across 11$\times$11 FoVs. The remaining two ImagingLens (\#2 and \#3) are designed in Zemax by experienced optical engineers who exerted their best efforts to optimize the performance, also starting from scratch. We use the default merit function (RMS spot radius, x+y relative to the centroid) with 16 rings and 12 arms, with thickness constraints to prevent negative values. The entrance pupil diameter is fixed, and the F-number controls first-order properties. All radii, aspheric coefficients, and thicknesses are set as variables. Detailed lens data are provided in the supplementary material.

\begin{table}[tp]
    \centering
    \caption{Performance metrics grouped by the number of lens elements. The ``TaskLens'' design is compared against three different ``ImagingLens'' designs within each group. The various evaluation metrics do not consistently yield the same results; specifically, an optical lens that achieves the highest image classification accuracy may not necessarily demonstrate the best RMS spot size performance. Through task-driven lens design, it is feasible to develop optical lenses with simpler structures (fewer lens elements) that achieve comparable or even superior image classification accuracy when compared to classical lens designs that prioritize minimizing optical aberrations.}
    \label{tab:tasklens}
    \renewcommand{\arraystretch}{0.95}
    \resizebox{\linewidth}{!}{
    \begin{tabular}{l l c c c}
    \toprule
    \textbf{Lens Structure} & \textbf{Lens Design} & \textbf{Accuracy (\%)} & \textbf{PSNR (dB)} & \textbf{Avg RMS Spot Size ($\mu$m)} \\
    \midrule
    \multirow{4}{*}{2-Element} & TaskLens & \textcolor{red}{70.08} & 19.46 & 27.27 \\
                               & ImagingLens \#1    & 65.63     & 22.43 & 13.61 \\
                               & ImagingLens \#2    & \textcolor{blue}{68.54}  & \textcolor{red}{23.65} & \textcolor{red}{9.85}  \\
                               & ImagingLens \#3    & 67.01     & \textcolor{blue}{22.90} & \textcolor{blue}{10.99} \\
    \midrule
    \multirow{4}{*}{3-Element} & TaskLens & \textcolor{red}{73.40} & \textcolor{red}{23.85} & 10.41 \\
                               & ImagingLens \#1            & \textcolor{blue}{70.04}  & 22.77 & \textcolor{blue}{9.32}  \\
                               & ImagingLens \#2            & 69.92                  & 23.14 & 14.75 \\
                               & ImagingLens \#3            & 68.52                  & \textcolor{blue}{23.51} & \textcolor{red}{8.53}  \\
    \midrule
    \multirow{4}{*}{4-Element} & TaskLens & \textcolor{red}{73.61} & 23.67 & \textcolor{blue}{9.97}  \\
                               & ImagingLens \#1            & \textcolor{blue}{72.27}  & \textcolor{red}{25.00} & \textcolor{red}{7.32}  \\
                               & ImagingLens \#2            & 68.88                  & 23.63 & 10.11 \\
                               & ImagingLens \#3            & 68.56                  & \textcolor{blue}{23.98} & 11.37 \\
    \bottomrule
    \end{tabular}
    }
\end{table}

Table~\ref{tab:tasklens} presents the average top-1 classification accuracy across nine evaluation FoVs for each lens. The TaskLens consistently outperforms the corresponding ImagingLens in terms of image classification accuracy when the number of lens elements is the same. Due to optical aberrations, real-world image classification performance falls short of the upper bound of 75.63\%, which is calculated on high-quality testing images. The ImagingLens experiences a more pronounced performance drop, attributed to the misalignment between the design objectives and the evaluation metrics. Notably, the two-element TaskLens surpasses all three-element ImagingLenses, while the three-element TaskLens outshines all four-element ImagingLenses. This suggests that task-driven lens design can identify simpler optical structures with competitive, or even superior, computer vision performance.

To elucidate the reason for the performance improvement, we analyze the spot diagrams and PSFs of the TaskLens and the best-performing ImagingLens. As illustrated in Figure~\ref{fig:detailed_comparison}, both designs exhibit noticeable blur due to aberrations, which results from the limited number of optical elements that can only partially correct optical aberrations. The distinction lies in their aberration-correction strategies: the TaskLens focuses the majority of light into a compact core while allowing a small fraction to form extended, low-energy tails. This results in a distinctive long-tailed PSF, particularly evident in the three- and four-element TaskLenses. While this long-tailed distribution redistributes optical energy and leads to reduced contrast (manifesting as a ``haze'' in images), it offers a critical advantage: a sharp central peak. In scenarios with strict hardware constraints, perfectly correcting all aberrations is infeasible. Traditional designs (minimizing RMS spot size) often result in a broadened central spot to minimize outlier rays, which suppresses high-frequency spatial information. In contrast, the TaskLens optimization allows for a long-tailed distribution (larger RMS spot size) to maintain a highly concentrated core. This sharp peak effectively preserves high-frequency structural details (such as edges) that are vital for feature extraction in computer vision models. Although these aberrations degrade standard image quality metrics, our experiments suggest that deep networks are robust to the global contrast reduction caused by the tail but are highly sensitive to the loss of high-frequency information, allowing the network to effectively learn to extract useful feature information during the task-driven design process.

In contrast, the ImagingLens, which is designed according to classical objectives and decoupled from the computer vision task, strives to focus all rays as tightly as possible. When residual optical aberrations are present, the classical lens design objectives become ineffective, lacking guidance regarding which image features are preferred by computer vision tasks. These findings suggest that for resource-constrained systems, such as camera lenses for robotics, a task-driven lens design objective that accommodates certain task-agnostic aberrations in favor of preserving task-critical features can result in superior end-to-end performance.

\begin{table}[t]
    \caption{Cross-task performance evaluation matrix. Each row represents a different computer vision task for design, while the columns indicate the computer vision tasks used for evaluation. Bold values indicate the best performance for each evaluation task.}
    \label{tab:task_matrix}
    \centering
    \renewcommand{\arraystretch}{1.25}
    \resizebox{\linewidth}{!}{
    \begin{tabular}{l|c|c|c|c|c}
    \toprule
    \diagbox{Design Task}{Evaluation Task} & \makecell{RMS spot size\\ $\mu$m} & \makecell{Image Classification \\ Acc. (\%)} & \makecell{Object Detection \\ mAP.} & \makecell{Semantic Segmentation \\ mAP} & \makecell{Image-Text Retrieval \\ Recall@1}\\
    \midrule
    RMS spot size           & \textbf{9.32}  & 70.04          & 35.45           & 36.08     & 54.4     \\
    Image Classification    & 10.41          & \cellcolor{orange!30}\textbf{73.40} & \cellcolor{orange!30}40.80           & \cellcolor{orange!30}41.44          & \cellcolor{orange!30}68.2 \\
    Object Detection        & 15.27          & \cellcolor{orange!30}72.18          & \cellcolor{orange!30}\textbf{42.56}  & \cellcolor{orange!30}40.21          & \cellcolor{orange!30}67.4 \\
    Semantic Segmentation   & 14.86          & \cellcolor{orange!30}72.60          & \cellcolor{orange!30}41.75           & \cellcolor{orange!30}\textbf{43.41} & \cellcolor{orange!30}65.5 \\
    Image-Text Retrieval    & 12.90          & \cellcolor{orange!30}71.21          & \cellcolor{orange!30}42.25           & \cellcolor{orange!30}42.32          & \cellcolor{orange!30}\textbf{70.4} \\
    \bottomrule
    \end{tabular}
    }
\end{table}

\subsection{Various downstream computer vision tasks}
\label{sec:downstream_tasks}

To assess the versatility of our design approach, we evaluate the lenses across a broader range of downstream applications, including object detection, semantic segmentation, and image-text retrieval in VLMs. These tasks effectively represent real-world challenges in computer vision. For object detection, we employ the Faster R-CNN model with a ResNet-50-FPN backbone from Detectron2~\cite{girshick2015fast,wu2019detectron2}, trained and tested on the COCO 2017 dataset~\cite{lin2014microsoft}. Performance is quantified using the mean Average Precision (mAP) metric, which evaluates a model's ability to accurately detect and classify objects. In semantic segmentation, we use the Mask2Former model with a Swin-Small backbone from HuggingFace~\cite{cheng2022masked}, also trained and tested on the COCO 2017 dataset. This model's performance is similarly quantified using the mAP metric, which assesses its ability to accurately segment objects. For image-text retrieval, we utilize the CLIP model with a ViT-Large-Patch14 backbone from HuggingFace~\cite{radford2021learning}, trained and tested on the Flickr30k dataset~\cite{young2014image}. In this context, optical lenses are trained to maximize the similarity between image encodings and text encodings via a contrastive loss. Performance is evaluated using the top-1 recall metric.

We employ the three-element ``TaskLens'' and ``ImagingLens'' (\#2) from the previous image classification section and design new lenses for the other three computer vision tasks, maintaining consistent experimental settings. The experimental results indicate that task-driven lens design is effective across all three tasks when designing lenses from scratch. We subsequently evaluate the performance of all these lenses across all metrics, with results presented in Table~\ref{tab:task_matrix}. Our findings reveal that training for a specific computer vision task invariably produces superior performance during evaluation. Moreover, the four computer vision tasks exhibit notable similarities. Lenses designed for one task perform well on others, significantly outperforming those designed with classical objectives. We believe this success stems from the need for accurate preservation of image textures and high-frequency features across these tasks, highlighting their generalizability. This finding further inspires the design of computer vision lenses by starting with simpler tasks (e.g., image classification) that can generalize to more complex tasks (e.g., VLM), which often present challenges for stable convergence. Example qualitative visualization examples for object detection and segmentation are provided in the supplementary material, demonstating the network outputs of the TaskLens and ImagingLens.

\subsection{Optical simulation validation}
\label{sec:simulation_validation}

To validate the fidelity of our optical simulation framework, we conducted experimental comparisons using a commercial camera system: a Canon EOS R6 paired with a Canon RF50mm f/1.8 lens. Experiments were performed in a controlled laboratory setting. We set the aperture to f/1.8 to maximize optical aberrations, thereby providing a more rigorous test of the simulator's capabilities. Both the focus distance and the object plane were set to 1.0 m. We employed two standard metrics for optical characterization: the Modulation Transfer Function (MTF) and the Point Spread Function (PSF). MTF quantifies the system's resolution and contrast transfer, while the PSF characterizes the impulse response and aberration profile. Data acquisition was performed in RAW format to avoid non-linear ISP processing. We used the ``rawpy'' library~\cite{rawpy} to linearize the images, ensuring that the pixel values remained proportional to the scene illuminance.

\begin{figure}[!htp]
    \centering
    \includegraphics[width=\linewidth]{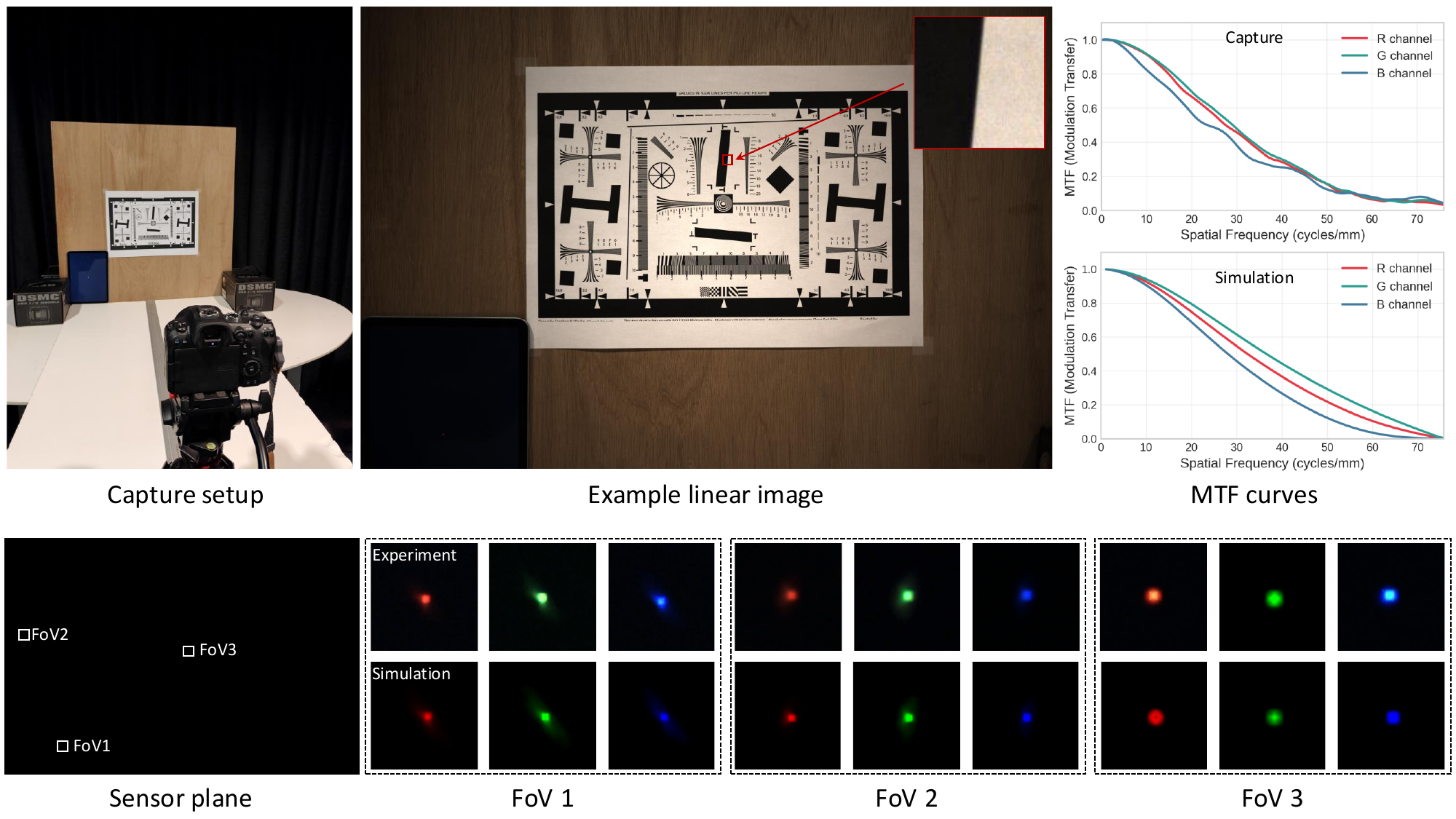}
    \caption{Validation of the optical simulation pipeline using a Canon EOS R6 and RF50mm f/1.8 lens. The experimental setup utilizes a test chart for MTF calculation and a monitor-displayed point source for PSF measurement. The measured MTF curves (top) track the simulated values closely, confirming accurate frequency response modeling. The captured PSFs (bottom) at three FoV exhibit strong morphological agreement with the simulation, validating the modeling of field-dependent aberrations.}
    \label{fig:psf_measurement}
\end{figure}

For MTF validation, we imaged a standard test chart. We extracted a region of interest (ROI) and computed the MTF following the methodology described in~\cite{chen2022computational}. We then simulated the same scene and optical parameters in our simulator. As shown in Fig.~\ref{fig:psf_measurement}, the measured MTF curves exhibit a strong correlation with the simulated results across the spatial frequency spectrum. Minor deviations are attributed to manufacturing tolerances of the physical lens and sensor noise; however, the overall trend validates the frequency response modeling of our simulator.

To validate the PSF, we measured the system's impulse response at three distinct fields of view (FoVs) to capture field-dependent variations. We generated a pseudo-point source by displaying a 2$\times$2 pixel bright spot on a high-resolution iPad display. While not an ideal point source, this approximation allows for practical empirical measurement. To ensure a fair comparison, the simulated PSFs were convolved with a 2$\times$2 pixel kernel matching the physical source dimensions. Figure~\ref{fig:psf_measurement} presents the comparison: the simulated PSFs closely reproduce the morphology and energy distribution of the captured PSFs. This strong agreement confirms the simulator's accuracy in modeling complex optical aberrations.

\section{Extra validation and robustness analyses}

We present additional validation experiments that assess the robustness and practical utility of the proposed task-driven lens design. Specifically, we analyze manufacturing tolerance, examine generalization across different network architectures, compare against conventional end-to-end optical co-design, and evaluate performance with and without post-capture image restoration.

\subsection{Lens manufacturing tolerance analysis}  
\label{sec:tolerance_analysis}  

Manufacturing and assembly imperfections are inevitable in practical applications and can significantly degrade the performance of optical systems, subsequently impacting downstream applications. Consequently, evaluating a lens design's tolerance to such errors is a critical component of the design process. To assess the robustness of our designs, we simulated the effects of manufacturing and assembly errors by generating three randomly ``perturbed'' versions of each lens. The classification performance of these perturbed lenses was then evaluated using a neural network model trained on the original, unperturbed design.

The results, as presented in Table~\ref{tab:manufacture_error}, illustrate the performance of each lens before and after the introduction of random errors. A comparative analysis of the average classification accuracy reveals that our TaskLens consistently outperforms the conventional ImagingLens, even in the presence of these perturbations. Notably, TaskLens designs exhibit superior robustness to manufacturing tolerances, demonstrating a smaller decrease in performance relative to their ImagingLens counterparts. A striking example is observed in the three-element comparison, where the ImagingLens experiences an average performance degradation of 3.77\%, whereas the three-element TaskLens only shows a minor decrease of 0.56\%. This enhanced tolerance to fabrication errors constitutes a significant practical advantage. We attribute this robustness to our task-driven design philosophy, which does not necessitate perfect optical aberration correction, thereby making it more tolerant of the minor, unintended aberrations introduced during manufacturing and assembly processes.

\begin{table*}[!htp]
    \centering
    \scriptsize
    \renewcommand{\arraystretch}{1.5}
    \caption{Image classification accuracy of different lenses with random manufacturing and assembly errors. From left to right: 2-element TaskLens, 2-element ImagingLens (\#2), 3-element TaskLens, 3-element ImagingLens (\#1), 4-element TaskLens, 4-element ImagingLens (\#1).}
    \resizebox{\textwidth}{!}{
    \begin{tabular}{l|cc|cc|cc}
    \toprule
    \multirow{2}{*}{Lens} & \multicolumn{2}{c}{2-Element} & \multicolumn{2}{c}{3-Element} & \multicolumn{2}{c}{4-Element}
    \\
    \cmidrule{2-7}
     & TaskLens & ImagingLens (\#2) & TaskLens & ImagingLens (\#1) & TaskLens & ImagingLens (\#1)
     \\
    \midrule
    Designed & 70.08\% & 68.54\% & 73.40\% & 70.04\% & 73.61\% & 72.27\% \\
    \midrule
    Lens \#1 & 69.73\% & 67.43\% & 72.82\% & 66.24\% & 72.58\% & 71.07\% \\
    Lens \#2 & 69.65\% & 67.27\% & 72.91\% & 66.36\% & 72.66\% & 70.99\% \\
    Lens \#3 & 69.83\% & 67.23\% & 72.78\% & 66.21\% & 72.65\% & 70.98\% \\
    \midrule
    Avg (\#1 $\sim$ \#3) & \textbf{69.74\%} & 67.31\% & \textbf{72.84\%} & 66.27\% & \textbf{72.63\%} & 71.10\% \\
    Avg decrease & \textbf{-0.34\%} & -1.23\% & \textbf{-0.56\%} & -3.77\% & \textbf{-0.98\%} & -1.17\% \\
    \bottomrule
    \end{tabular}
    }
    \\[5pt]
    \begin{flushleft}
        \scriptsize
        \textbf{Designed}: The ideal lens design without manufacturing or assembly errors. \\
        \textbf{Lens \#1-3}: Three instances of the designed lens with random manufacturing and assembly errors. \\
    \end{flushleft}
    \label{tab:manufacture_error}
\end{table*}

\subsection{Performance generalization across different network architectures}
\label{sec:generalization}

In our task-driven lens design, the lens is optimized specifically for a particular network architecture, namely ResNet-50. However, in practical applications, the downstream network may vary. For instance, smaller models might be deployed on edge devices with limited computational resources, while larger models may be employed in cloud environments to enhance performance. Since optical lenses cannot be modified post-manufacture, it is essential to ascertain whether the performance advantages of a TaskLens can be preserved when integrated with different network architectures.

To investigate this, we evaluated our lenses using three additional network models that vary significantly in size and architecture: MobileNetV3-Large~\cite{howard2017mobilenets} (``MobileNetV3-L''), a 5.5M-parameter CNN; Swin Transformer-Base~\cite{liu2022swin} (``Swin-B''), an 88M-parameter Transformer; and Vision Transformer-Large~\cite{dosovitskiy2020image} (``ViT-L/16''), a 304M-parameter Transformer. These are compared against the original 26M-parameter ResNet-50, which is also a CNN architecture. The image classification results are summarized in Table~\ref{tab:generalization}. Across all tested network models, our TaskLens consistently achieves the highest accuracy for each specified number of lens elements. This finding demonstrates that the TaskLens is compatible with various network architectures and remains robustly advantageous in terms of performance. Moreover, this result bears practical implications: one can utilize a smaller, computationally efficient network during the lens optimization phase, thereby reducing design time and resource requirements, and subsequently deploy the optimized lens with a larger, more powerful network to maximize overall performance.

\begin{table}[!htp]
    \caption{Classification accuracy with different network models. The designed TaskLenses are compatible with different network models while maintaining enhanced performance compared to the corresponding ImagingLenses. The results for the original ResNet50 are reported in Table~\ref{tab:tasklens}. Red/blue highlighting indicates best/second-best performance within each configuration. These results support the claim that task-driven lens design can identify optical characteristics that are compatible with different network architectures.}
    \label{tab:generalization}
    \centering
    \scriptsize
    \renewcommand{\arraystretch}{0.9}
    \resizebox{\textwidth}{!}{
    \begin{tabular}{l|l|c|c|c}
    \toprule
    \multirow{2}{*}{\textbf{Lens Elements}} & \multirow{2}{*}{\textbf{Method}} & \textbf{MobileNetV3-L}~\cite{howard2019searching} & \textbf{Swin-B}~\cite{liu2022swin} & \textbf{ViT-L/16}~\cite{dosovitskiy2020image} \\
     & & (5.4M params) & (88M params) & (304M params) \\
    \midrule
    \multirow{4}{*}{2-Element} 
    & TaskLens (Ours) & \textcolor{red}{68.22\%} & \textcolor{red}{81.19\%} & \textcolor{red}{81.76\%} \\
    & ImagingLens \#1 & 64.60\% & 79.03\% & 79.39\% \\
    & ImagingLens \#2 & \textcolor{blue}{67.73\%} & \textcolor{blue}{80.87\%} & \textcolor{blue}{81.19\%} \\
    & ImagingLens \#3 & 66.05\% & 80.23\% & 80.79\% \\
    \midrule
    \multirow{4}{*}{3-Element} 
    & TaskLens (Ours) & \textcolor{red}{71.82\%} & \textcolor{red}{82.65\%} & \textcolor{red}{83.46\%} \\
    & ImagingLens \#1 & \textcolor{blue}{68.36\%} & \textcolor{blue}{81.19\%} & \textcolor{blue}{81.62\%} \\
    & ImagingLens \#2 & 67.94\% & 81.08\% & 81.46\% \\
    & ImagingLens \#3 & 68.01\% & 80.19\% & 81.61\% \\
    \midrule
    \multirow{4}{*}{4-Element} 
    & TaskLens (Ours) & \textcolor{red}{72.06\%} & \textcolor{red}{82.82\%} & \textcolor{red}{83.62\%} \\
    & ImagingLens \#1 & \textcolor{blue}{71.00\%} & \textcolor{blue}{82.43\%} & \textcolor{blue}{82.52\%} \\
    & ImagingLens \#2 & 66.78\% & 80.75\% & 81.60\% \\
    & ImagingLens \#3 & 67.19\% & 80.46\% & 81.50\% \\
    \bottomrule
    \end{tabular}
    }
\end{table}

\subsection{Comparison with end-to-end optical design from successful starting points}
\label{sec:end2end_comparison}

In this section, we investigate whether an end-to-end optical design approach can identify solutions comparable to our TaskLens. To test this hypothesis, we conducted a joint optimization of both the optical system and a pre-trained ResNet-50 network, which is a standard methodology for end-to-end design. We evaluated two distinct starting points for the lens: (1) a randomly initialized optical lens, representing a design from scratch, and (2) our best-performing, pre-optimized ImagingLens. The final classification accuracies are summarized in Table~\ref{tab:end2end}. Our results indicate that the end-to-end training process fails to converge when initiated from an all-flat optics configuration. This phenomenon can be attributed to the highly non-convex optimization landscape inherent to optical design, which makes convergence from a random or null starting point particularly challenging.

When the training is initialized with a well-designed ImagingLens, the end-to-end optimization achieves only marginal performance improvements and fails to reach the accuracy of our TaskLens. We hypothesize that this occurs because the optimization becomes trapped in a strong local minimum. The initial ImagingLens configuration is already highly optimized for traditional objectives. Consequently, the backpropagated gradients are typically insufficient to move the design out of this local minimum towards a more globally optimal solution discovered by our method. As a result, the end-to-end process primarily fine-tunes the initial lens design, yielding classification performance that remains inferior to that of our TaskLens. This experiment underscores the limitations of conventional end-to-end fine-tuning and highlights the effectiveness of our design-space exploration in uncovering novel, task-optimal optical configurations.

\begin{table}[!htp]
    \caption{End-to-end design from scratch fails to converge. When starting with a well-designed ImagingLens, the end-to-end process fails to discover the optimal classification lens.}
    \label{tab:end2end}
    \centering
    \renewcommand{\arraystretch}{0.8}
    \resizebox{0.7\linewidth}{!}{
    \begin{tabular}{l|c|c|c}
        \toprule
        & \multirow{2}{*}{TaskLens} & \multicolumn{2}{c}{End-to-End Training}  \\
        \cmidrule{3-4}
        & & From ImagingLens & From Scratch \\
        \midrule
        2-Element & 70.05\% & 69.55\%  & \redcross \\
        \midrule
        3-Element & 73.40\% & 71.94\%  & \redcross  \\
        \midrule
        4-Element & 73.61\% & 73.44\%  & \redcross \\
        \bottomrule
    \end{tabular}
    }
\end{table}

\subsection{Effect of post-capture image restoration}
\label{sec:image_restoration}

Image restoration is a widely employed technique for mitigating optical aberrations in captured images. Modern algorithms, particularly those based on neural networks~\cite{chen2022computational, yang2025efficient}, have demonstrated effectiveness in recovering fine image details, thereby enhancing the performance of downstream vision tasks. This raises a critical question: Can post-capture image restoration bridge the performance gap between our TaskLenses and conventional ImagingLenses? 

To address this question, we applied an image restoration process to the outputs of all lenses and compared the resulting classification accuracies. For image restoration, we utilized NAFNet~\cite{chen2022simple}, a state-of-the-art restoration model, and we trained a separate NAFNet for each lens on its simulated images. The results are summarized in Table~\ref{tab:image_restoration}. While image restoration consistently enhances classification accuracy across all lenses, the TaskLens continues to outperform the corresponding ImagingLens. This finding suggests that the performance advantage of the TaskLens is not solely attributable to a type of blur that can be easily corrected by an image restoration algorithm. Instead, the task-specific optical encoding provides a persistent advantage, indicating that the gap in classification performance cannot be bridged by image restoration alone.

\begin{table}[!htp]
    \caption{Classification performance after image restoration. Improvements are observed in both PSNR and classification accuracy for all designs. Despite the TaskLens do not demonstrate the best PSNR scores, they maintain the best image classification performance. This indicates that the image classification advantages stem from fundamental optical properties rather than aberration tolerance. Red/blue highlighting indicates best/second-best performance within each configuration.}
    \label{tab:image_restoration}
    \centering
    \scriptsize
    \renewcommand{\arraystretch}{0.9}
    \resizebox{0.8\linewidth}{!}{
    \begin{tabular}{l|l|c|c}
        \toprule
        \multirow{2}{*}{\textbf{Lens Elements}} & \multirow{2}{*}{\textbf{Method}} & \textbf{PSNR [dB]} & \textbf{Classification Accuracy} \\
         & & & \\
        \midrule
        \multirow{4}{*}{2-Element} 
        & TaskLens (Ours) & 27.24 & \textcolor{red}{72.03\%} \\
        & ImagingLens \#1 & 27.54 & 68.19\% \\
        & ImagingLens \#2 & \textcolor{blue}{30.30} & \textcolor{blue}{71.24\%} \\
        & ImagingLens \#3 & \textcolor{red}{30.87} & 71.08\% \\
        \midrule
        \multirow{4}{*}{3-Element} 
        & TaskLens (Ours) & \textcolor{red}{32.31} & \textcolor{red}{74.43\%} \\
        & ImagingLens \#1 & 29.44 & 72.42\% \\
        & ImagingLens \#2 & \textcolor{blue}{30.47} & \textcolor{blue}{73.35\%} \\
        & ImagingLens \#3 & 29.95 & 70.90\% \\
        \midrule
        \multirow{4}{*}{4-Element} 
        & TaskLens (Ours) & \textcolor{blue}{33.58} & \textcolor{red}{74.61\%} \\
        & ImagingLens \#1 & \textcolor{red}{34.29} & \textcolor{blue}{73.98\%} \\
        & ImagingLens \#2 & 29.69 & 72.16\% \\
        & ImagingLens \#3 & 29.38 & 71.52\% \\
        \bottomrule
    \end{tabular}
    }
\end{table}

\section{Discussion and conclusion}

\paragraph{Application scenarios} In applications where high-quality, perfectly imaging lenses are infeasible due to form-factor and cost constraints (e.g., robotics), residual aberrations can significantly affect computer vision performance. In these scenarios, simple lenses designed with our task-driven objective bridge the gap between optical design and downstream tasks. This approach better preserves features preferred by computer vision models, improving performance despite aberrations. Furthermore, while fine-tuning vision models on corrupted images is a common approach to enhance performance, it poses extra computational cost and may be unaffordable in some cases as modern network architectures become more complex and training datasets grow larger. In scenarios where users directly deploy pre-trained models, our task-driven lens design objective offers a valuable alternative for designing lenses that are inherently more compatible with computer vision models than those designed with traditional objectives.

\paragraph{Challenges and future work} Our experiments reveal that as the complexity of computer vision models increases, the gradients provided by the networks become more unstable, leading to oscillatory optimization behavior. For instance, when training with the BLIP model~\cite{li2022blip} for image-text retrieval, we encountered significant convergence difficulties with our task-driven lens design. We believe this struggle arises from the noise in gradients produced by the complex network architecture during backpropagation. Moreover, propagating gradients through the entire network poses challenges related to GPU memory consumption. These issues not only complicate direct optimization for certain modern computer vision tasks and network architectures but also motivate us to explore more general evaluative objectives for the feature encoding capabilities of optical lenses. Consequently, we are currently undertaking follow-up work to simplify the task-driven lens design objective while elucidating key underlying principles for task-driven lens design.

\paragraph{Conclusion} In this work, we introduce task-driven lens design. By leveraging a well-trained neural network to inform optical optimization, we demonstrate the feasibility of designing simpler yet more effective lenses from scratch for computer vision applications. When lens structures alone are insufficient to correct all optical aberrations, our resulting ``TaskLenses'' consistently outperform those created using classical design objectives. The unique optical characteristics and PSF effectively preserve critical image features identified by vision foundation models, and the TaskLenses exhibit greater tolerance for manufacturing errors. The generalizability across various computer vision tasks and network architectures encourages optimization using simpler tasks and networks, with the potential to scale to more complex tasks and networks. We believe that task-driven lens design presents a new paradigm for the next generation of computational camera lenses, particularly for optical lenses constrained by practical considerations such as form factor.





\begin{backmatter}
\bmsection{Funding}
King Abdullah University of Science and Technology (Individual Baseline Funding); King Abdullah University of Science and Technology (Center of Excellence for Generative AI); Fonds Wetenschappelijk Onderzoek (G0A3O24N); Vrije Universiteit Brussel (Hercules, Methusalem, OZR).


\bmsection{Disclosures}
The authors declare no conflicts of interest.

\bmsection{Data availability} Data underlying the results presented in this paper are not publicly available at this time but may be obtained from the authors upon reasonable request.

\bmsection{Supplemental document} See Supplement 1 for supporting content.

\end{backmatter}



\bibliography{sample}

@String(CVPR= {IEEE Conf. Comput. Vis. Pattern Recog.})

@String(ICCV= {Int. Conf. Comput. Vis.})

@String(ECCV= {Eur. Conf. Comput. Vis.})

@String(TOG= {ACM Trans. Graph.})

@String(CVPR  = {CVPR})

@String(ICCV  = {ICCV})

@String(ECCV  = {ECCV})

@String(TOG   = {ACM Trans. Graph.})

@String(OL = {Opt. Lett.})

@String(OE = {Opt. Express})

@String(ICCP= {Int. Conf. Comput. Photog.})

@inproceedings{deng2009imagenet,
  title={{ImageNet}: A large-scale hierarchical image database},
  author={Deng, Jia and Dong, Wei and Socher, Richard and Li, Li-Jia and Li, Kai and Fei-Fei, Li},
  booktitle=CVPR,
  pages={248--255},
  year={2009},
  organization={Ieee}
}

@article{young2014image,
  title={From image descriptions to visual denotations: New similarity metrics for semantic inference over event descriptions},
  author={Young, Peter and Lai, Alice and Hodosh, Micah and Hockenmaier, Julia},
  journal={Transactions of the association for computational linguistics},
  volume={2},
  pages={67--78},
  year={2014},
  publisher={MIT Press One Rogers Street, Cambridge, MA 02142-1209, USA journals-info~…}
}

@inproceedings{lin2014microsoft,
  title={Microsoft coco: Common objects in context},
  author={Lin, Tsung-Yi and Maire, Michael and Belongie, Serge and Hays, James and Perona, Pietro and Ramanan, Deva and Doll{\'a}r, Piotr and Zitnick, C Lawrence},
  booktitle={Computer vision--ECCV 2014: 13th European conference, zurich, Switzerland, September 6-12, 2014, proceedings, part v 13},
  pages={740--755},
  year={2014},
  organization={Springer}
}

@inproceedings{girshick2015fast,
  title={Fast r-cnn},
  author={Girshick, Ross},
  booktitle={Proceedings of the IEEE international conference on computer vision},
  pages={1440--1448},
  year={2015}
}

@misc{sugita2015zoom,
  title={Zoom lens and image pickup apparatus including same},
  author={Sugita, Shigenobu},
  year={2015},
  month=aug # "~18",
  publisher={Google Patents},
  note={US Patent 9,110,278}
}

@inproceedings{he2016deep,
  title={Deep residual learning for image recognition},
  author={He, Kaiming and Zhang, Xiangyu and Ren, Shaoqing and Sun, Jian},
  booktitle=CVPR,
  pages={770--778},
  year={2016}
}

@inproceedings{redmon2016you,
  title={You only look once: Unified, real-time object detection},
  author={Redmon, Joseph and Divvala, Santosh and Girshick, Ross and Farhadi, Ali},
  booktitle=CVPR,
  pages={779--788},
  year={2016}
}

@article{loshchilov2017decoupled,
  title={Decoupled weight decay regularization},
  author={Loshchilov, Ilya and Hutter, Frank},
  journal={arXiv preprint arXiv:1711.05101},
  year={2017}
}

@article{howard2017mobilenets,
  title={{MobileNets}: Efficient convolutional neural networks for mobile vision applications},
  author={Howard, Andrew G and Zhu, Menglong and Chen, Bo and Kalenichenko, Dmitry and Wang, Weijun and Weyand, Tobias and Andreetto, Marco and Adam, Hartwig},
  journal={arXiv preprint arXiv:1704.04861},
  year={2017}
}

@article{sitzmann2018end,
  title={End-to-end optimization of optics and image processing for achromatic extended depth of field and super-resolution imaging},
  author={Sitzmann, Vincent and Diamond, Steven and Peng, Yifan and Dun, Xiong and Boyd, Stephen and Heidrich, Wolfgang and Heide, Felix and Wetzstein, Gordon},
  journal=TOG,
  volume={37},
  number={4},
  pages={1--13},
  year={2018},
  publisher={ACM New York, NY, USA}
}

@misc{abe2018zoom,
  title={Zoom lens and image pickup apparatus including the same},
  author={Abe, Hirofumi},
  year={2018},
  month=apr # "~24",
  publisher={Google Patents},
  note={US Patent 9,952,446}
}

@inproceedings{howard2019searching,
  title={Searching for {MobileNetv3}},
  author={Howard, Andrew and Sandler, Mark and Chu, Grace and Chen, Liang-Chieh and Chen, Bo and Tan, Mingxing and Wang, Weijun and Zhu, Yukun and Pang, Ruoming and Vasudevan, Vijay and others},
  booktitle=ICCV,
  pages={1314--1324},
  year={2019}
}

@inproceedings{chang2019deep,
  title={Deep optics for monocular depth estimation and {3D} object detection},
  author={Chang, Julie and Wetzstein, Gordon},
  booktitle=ICCV,
  pages={10193--10202},
  year={2019}
}

@article{jeon2019compact,
  title={Compact snapshot hyperspectral imaging with diffracted rotation},
  author={Jeon, Daniel S and Baek, Seung-Hwan and Yi, Shinyoung and Fu, Qiang and Dun, Xiong and Heidrich, Wolfgang and Kim, Min H},
  journal=TOG,
  volume={38},
  number={4},
  pages={1--13},
  year={2019},
  publisher={ACM New York, NY, USA}
}

@misc{wu2019detectron2,
  author =       {Yuxin Wu and Alexander Kirillov and Francisco Massa and
                  Wan-Yen Lo and Ross Girshick},
  title =        {Detectron2},
  howpublished = {\url{https://github.com/facebookresearch/detectron2}},
  year =         {2019}
}

@misc{rawpy,
  title =        {rawpy},
  author =       {{rawpy contributors}},
  howpublished = {[Software] \url{https://github.com/letmaik/rawpy} (accessed 22 Jan. 2026)},
  year =         {2026}
}

@article{dosovitskiy2020image,
  title={An image is worth 16x16 words: {Transformers} for image recognition at scale},
  author={Dosovitskiy, Alexey and Beyer, Lucas and Kolesnikov, Alexander and Weissenborn, Dirk and Zhai, Xiaohua and Unterthiner, Thomas and Dehghani, Mostafa and Minderer, Matthias and Heigold, Georg and Gelly, Sylvain and others},
  journal={arXiv preprint arXiv:2010.11929},
  year={2020}
}

@inproceedings{metzler2020deep,
  title={Deep optics for single-shot high-dynamic-range imaging},
  author={Metzler, Christopher A and Ikoma, Hayato and Peng, Yifan and Wetzstein, Gordon},
  booktitle=CVPR,
  pages={1375--1385},
  year={2020}
}

@article{dun2020learned,
  title={Learned rotationally symmetric diffractive achromat for full-spectrum computational imaging},
  author={Dun, Xiong and Ikoma, Hayato and Wetzstein, Gordon and Wang, Zhanshan and Cheng, Xinbin and Peng, Yifan},
  journal={Optica},
  volume={7},
  number={8},
  pages={913--922},
  year={2020},
  publisher={Optical Society of America}
}

@misc{shabtay2020folded,
  title={Folded camera lens designs},
  author={Shabtay, Gal and Goldenberg, Ephraim and Dror, Michael and Yedid, Itay and Bachar, Gil},
  year={2020},
  month=feb # "~25",
  publisher={Google Patents},
  note={US Patent 10,571,644}
}

@inproceedings{sun2020learning,
  title={Learning {Rank-1} diffractive optics for single-shot high dynamic range imaging},
  author={Sun, Qilin and Tseng, Ethan and Fu, Qiang and Heidrich, Wolfgang and Heide, Felix},
  booktitle=CVPR,
  pages={1386--1396},
  year={2020}
}

@inproceedings{baek2021single,
  title={Single-shot hyperspectral-depth imaging with learned diffractive optics},
  author={Baek, Seung-Hwan and Ikoma, Hayato and Jeon, Daniel S and Li, Yuqi and Heidrich, Wolfgang and Wetzstein, Gordon and Kim, Min H},
  booktitle=ICCV,
  pages={2651--2660},
  year={2021}
}

@article{cote2021deep,
  title={Deep learning-enabled framework for automatic lens design starting point generation},
  author={C{\^o}t{\'e}, Geoffroi and Lalonde, Jean-Fran{\c{c}}ois and Thibault, Simon},
  journal=OE,
  volume={29},
  number={3},
  pages={3841--3854},
  year={2021},
  publisher={Optica Publishing Group}
}

@inproceedings{radford2021learning,
  title={Learning transferable visual models from natural language supervision},
  author={Radford, Alec and Kim, Jong Wook and Hallacy, Chris and Ramesh, Aditya and Goh, Gabriel and Agarwal, Sandhini and Sastry, Girish and Askell, Amanda and Mishkin, Pamela and Clark, Jack and others},
  booktitle={International conference on machine learning},
  pages={8748--8763},
  year={2021},
  organization={PmLR}
}

@inproceedings{ikoma2021depth,
  title={Depth from defocus with learned optics for imaging and occlusion-aware depth estimation},
  author={Ikoma, Hayato and Nguyen, Cindy M and Metzler, Christopher A and Peng, Yifan and Wetzstein, Gordon},
  booktitle=ICCP,
  pages={1--12},
  year={2021},
  organization={IEEE}
}

@article{sun2021end,
  title={End-to-end complex lens design with differentiable ray tracing},
  author={Sun, Qilin and Wang, Congli and Qiang, Fu and Xiong, Dun and Wolfgang, Heidrich},
  journal=TOG,
  volume={40},
  number={4},
  pages={1--13},
  year={2021}
}

@article{tseng2021neural,
  title={Neural nano-optics for high-quality thin lens imaging},
  author={Tseng, Ethan and Colburn, Shane and Whitehead, James and Huang, Luocheng and Baek, Seung-Hwan and Majumdar, Arka and Heide, Felix},
  journal={Nature Communications},
  volume={12},
  number={1},
  pages={6493},
  year={2021},
  publisher={Nature Publishing Group UK London}
}

@article{tseng2021differentiable,
  title={Differentiable compound optics and processing pipeline optimization for end-to-end camera design},
  author={Tseng, Ethan and Mosleh, Ali and Mannan, Fahim and St-Arnaud, Karl and Sharma, Avinash and Peng, Yifan and Braun, Alexander and Nowrouzezahrai, Derek and Lalonde, Jean-Francois and Heide, Felix},
  journal=TOG,
  volume={40},
  number={2},
  pages={1--19},
  year={2021},
  publisher={ACM New York, NY}
}

@inproceedings{liu2021swin,
  title={Swin transformer: Hierarchical vision transformer using shifted windows},
  author={Liu, Ze and Lin, Yutong and Cao, Yue and Hu, Han and Wei, Yixuan and Zhang, Zheng and Lin, Stephen and Guo, Baining},
  booktitle=ICCV,
  pages={10012--10022},
  year={2021}
}

@article{liu2021end,
  title={End-to-end computational optics with a singlet lens for large depth-of-field imaging},
  author={Liu, Yuankun and Zhang, Chongyang and Kou, Tingdong and Li, Yueyang and Shen, Junfei},
  journal=OE,
  volume={29},
  number={18},
  pages={28530--28548},
  year={2021},
  publisher={Optica Publishing Group}
}

@inproceedings{zhai2022scaling,
  title={Scaling vision transformers},
  author={Zhai, Xiaohua and Kolesnikov, Alexander and Houlsby, Neil and Beyer, Lucas},
  booktitle=CVPR,
  pages={12104--12113},
  year={2022}
}

@article{zhang2022dino,
  title={{DINO: DETR} with improved denoising anchor boxes for end-to-end object detection},
  author={Zhang, Hao and Li, Feng and Liu, Shilong and Zhang, Lei and Su, Hang and Zhu, Jun and Ni, Lionel M and Shum, Heung-Yeung},
  journal={arXiv preprint arXiv:2203.03605},
  year={2022}
}

@inproceedings{liu2022swin,
  title={Swin transformer v2: Scaling up capacity and resolution},
  author={Liu, Ze and Hu, Han and Lin, Yutong and Yao, Zhuliang and Xie, Zhenda and Wei, Yixuan and Ning, Jia and Cao, Yue and Zhang, Zheng and Dong, Li and others},
  booktitle=CVPR,
  pages={12009--12019},
  year={2022}
}

@inproceedings{cheng2022masked,
  title={Masked-attention mask transformer for universal image segmentation},
  author={Cheng, Bowen and Misra, Ishan and Schwing, Alexander G and Kirillov, Alexander and Girdhar, Rohit},
  booktitle={Proceedings of the IEEE/CVF conference on computer vision and pattern recognition},
  pages={1290--1299},
  year={2022}
}

@article{li2021end,
  title={End-to-end learned single lens design using fast differentiable ray tracing},
  author={Li, Zongling and Hou, Qingyu and Wang, Zhipeng and Tan, Fanjiao and Liu, Jin and Zhang, Wei},
  journal=OL,
  volume={46},
  number={21},
  pages={5453--5456},
  year={2021},
  publisher={Optica Publishing Group}
}

@inproceedings{li2022quantization,
  title={Quantization-aware deep optics for diffractive snapshot hyperspectral imaging},
  author={Li, Lingen and Wang, Lizhi and Song, Weitao and Zhang, Lei and Xiong, Zhiwei and Huang, Hua},
  booktitle=CVPR,
  pages={19780--19789},
  year={2022}
}

@inproceedings{li2022blip,
  title={Blip: Bootstrapping language-image pre-training for unified vision-language understanding and generation},
  author={Li, Junnan and Li, Dongxu and Xiong, Caiming and Hoi, Steven},
  booktitle={International conference on machine learning},
  pages={12888--12900},
  year={2022},
  organization={PMLR}
}

@incollection{chen2022simple,
  title={Simple Baselines for Image Restoration},
  author={Chen, Liangyu and Chu, Xiaojie and Zhang, Xiangyu and Sun, Jian},
  booktitle={Computer Vision -- ECCV 2022},
  publisher={Springer Nature Switzerland},
  year={2022},
  pages={17--33},
  doi={10.1007/978-3-031-20071-7_2}
}

@article{wang2022differentiable, title={dO: A Differentiable Engine for Deep Lens Design of Computational Imaging Systems}, volume={8}, ISSN={2573-0436}, url={http://dx.doi.org/10.1109/tci.2022.3212837}, DOI={10.1109/tci.2022.3212837}, journal={IEEE Transactions on Computational Imaging}, publisher={Institute of Electrical and Electronics Engineers (IEEE)}, author={Wang, Congli and Chen, Ni and Heidrich, Wolfgang}, year={2022}, pages={905–916} }

@article{chen2022computational, title={Computational Optics for Mobile Terminals in Mass Production}, volume={45}, ISSN={1939-3539}, url={http://dx.doi.org/10.1109/tpami.2022.3200725}, DOI={10.1109/tpami.2022.3200725}, number={4}, journal={IEEE Transactions on Pattern Analysis and Machine Intelligence}, publisher={Institute of Electrical and Electronics Engineers (IEEE)}, author={Chen, Shiqi and Lin, Ting and Feng, Huajun and Xu, Zhihai and Li, Qi and Chen, Yueting}, year={2023}, month=apr, pages={4245–4259} }

@inproceedings{cote2022differentiable,
  title={The differentiable lens: Compound lens search over glass surfaces and materials for object detection},
  author={C{\^o}t{\'e}, Geoffroi and Mannan, Fahim and Thibault, Simon and Lalonde, Jean-Fran{\c{c}}ois and Heide, Felix},
  booktitle=CVPR,
  pages={20803--20812},
  year={2023}
}

@article{yang2023curriculum, title={Curriculum learning for ab initio deep learned refractive optics}, volume={15}, ISSN={2041-1723}, url={http://dx.doi.org/10.1038/s41467-024-50835-7}, DOI={10.1038/s41467-024-50835-7}, number={1}, journal={Nature Communications}, publisher={Springer Science and Business Media LLC}, author={Yang, Xinge and Fu, Qiang and Heidrich, Wolfgang}, year={2024}, month=aug, pages={50835} }

@misc{zhou2023camera,
  title={Camera optical lens},
  author={Zhou, Xuepeng},
  year={2023},
  month=jan # "~31",
  publisher={Google Patents},
  note={US Patent 11,567,301}
}

@inproceedings{na2024end, series={SA ’24}, title={End-to-end Optimization of Fluidic Lenses}, url={http://dx.doi.org/10.1145/3680528.3687584}, DOI={10.1145/3680528.3687584}, booktitle={SIGGRAPH Asia 2024 Conference Papers}, publisher={ACM}, author={Na, Mulun and Jimenez Romero, Hector A. and Yang, Xinge and Klein, Jonathan and Michels, Dominik L. and Heidrich, Wolfgang}, year={2024}, month=dec, pages={1–10}, collection={SA ’24} }

@inproceedings{yang2024end, series={SA ’24}, title={End-to-End Hybrid Refractive-Diffractive Lens Design with Differentiable Ray-Wave Model}, url={http://dx.doi.org/10.1145/3680528.3687640}, DOI={10.1145/3680528.3687640}, booktitle={SIGGRAPH Asia 2024 Conference Papers}, publisher={ACM}, author={Yang, Xinge and Souza, Matheus and Wang, Kunyi and Chakravarthula, Praneeth and Fu, Qiang and Heidrich, Wolfgang}, year={2024}, month=dec, pages={1–11}, collection={SA ’24} }

@article{zhang2024vision,
  title={Vision-language models for vision tasks: A survey},
  author={Zhang, Jingyi and Huang, Jiaxing and Jin, Sheng and Lu, Shijian},
  journal={IEEE transactions on pattern analysis and machine intelligence},
  volume={46},
  number={8},
  pages={5625--5644},
  year={2024},
  publisher={IEEE}
}

@article{jiang2024minimalist, title={Minimalist and High-Quality Panoramic Imaging With PSF-Aware Transformers}, volume={33}, ISSN={1941-0042}, url={http://dx.doi.org/10.1109/tip.2024.3441370}, DOI={10.1109/tip.2024.3441370}, journal={IEEE Transactions on Image Processing}, publisher={Institute of Electrical and Electronics Engineers (IEEE)}, author={Jiang, Qi and Gao, Shaohua and Gao, Yao and Yang, Kailun and Yi, Zhonghua and Shi, Hao and Sun, Lei and Wang, Kaiwei}, year={2024}, pages={4568–4583} }

@article{yang2025efficient,
  title={Efficient Depth-and Spatially-Varying Image Simulation for Defocus Deblur},
  author={Yang, Xinge and Nguyen, Chuong and Wang, Wenbin and Kang, Kaizhang and Heidrich, Wolfgang and Li, Xiaoxing},
  journal={arXiv preprint arXiv:2507.00372},
  year={2025}
}

@article{teh2025automated,
  title={Automated design of compound lenses with discrete-continuous optimization},
  author={Teh, Arjun and Vicini, Delio and Bickel, Bernd and Gkioulekas, Ioannis and O'Toole, Matthew},
  journal={arXiv preprint arXiv:2509.23572},
  year={2025}
}






\end{document}